\documentclass[11pt,a4paper]{article}
\usepackage{microtype}

\usepackage[hyperref]{emnlp2018}
\usepackage[T2A,T1]{fontenc}
\usepackage[utf8]{inputenc}
\usepackage[russian,english]{babel}
\usepackage{multirow}

\usepackage{times}
\usepackage{latexsym}
\usepackage{amsmath}
\usepackage{amsfonts}
\usepackage{microtype}
\usepackage{xfrac}
\usepackage{url}
\usepackage{mathtools}
\usepackage{bbm}
\usepackage{tikz}
\usepackage[small]{caption}
\usetikzlibrary{shapes.geometric}
\usetikzlibrary{automata, positioning, arrows}
\usepackage{kotex}
\usepackage{booktabs}
\usepackage{adjustbox}
\usepackage{xspace}
\usepackage[normalem]{ulem}
\usepackage{subfig}
\usepackage{pifont}
\usepackage{xcolor}

\definecolor{awesome}{rgb}{1.0, 0.13, 0.32}
\definecolor{blue(pigment)}{rgb}{0.2, 0.2, 0.6}
\definecolor{darkcyan}{rgb}{0.0, 0.55, 0.55}
\definecolor{electricviolet}{rgb}{0.56, 0.0, 1.0}

\definecolor{cbcol1}{rgb}{1.00,0.75,0.75}
\definecolor{cbcol2}{rgb}{0.75,1.00,0.75}
\definecolor{cbcol3}{rgb}{0.75,0.75,1.00}
\definecolor{cbcol4}{rgb}{0.75,0.50,1.0}
\definecolor{cbcol5}{rgb}{1.00,0.75,0.0}
\definecolor{cbcol6}{rgb}{0.9,0.9,0.9}

\newcommand{\eos}{\textsc{eos}}
\newcommand*\circled[1]{
  \protect\tikz[baseline={([yshift=0.0pt]n.base)}]%
  \protect\node[shape=circle,inner sep=1pt,draw](n){\small #1};}

\newcommand*{\numberingBlueB}[1]{%
  \protect\tikz[baseline={([yshift=0.0pt]n.base)}]%
  \protect\node[fill=cbcol3,shape=circle,inner sep=1pt,draw](n){\small #1};}

\newcommand*{\numberingRedB}[1]{%
  \protect\tikz[baseline={([yshift=0.0pt]n.base)}]%
  \protect\node[fill=cbcol1 ,shape=circle,inner sep=1pt,draw](n){\small #1};}

\newcommand*{\numberingGreenB}[1]{%
  \protect\tikz[baseline={([yshift=0.0pt]n.base)}]%
  \protect\node[fill=cbcol2,shape=circle,inner sep=1pt,draw](n){\small #1};}

\newcommand*{\numberingYellowB}[1]{%
  \protect\tikz[baseline={([yshift=0.0pt]n.base)}]%
  \protect\node[fill=yellow!25,shape=circle,inner sep=1pt,draw](n){\small #1};}

\newcommand*{\numberingPurpleB}[1]{%
  \protect\tikz[baseline={([yshift=0.0pt]n.base)}]%
  \protect\node[fill=cbcol4,shape=rectangle,inner sep=2pt,draw](n){\small #1};}

\newcommand*{\numberingOrangeB}[1]{%
  \protect\tikz[baseline={([yshift=0.0pt]n.base)}]%
  \protect\node[fill=cbcol5,shape=rectangle,inner sep=2pt,draw](n){\small #1};}

\newcommand*{\numberingGrayB}[1]{%
  \protect\tikz[baseline={([yshift=0.0pt]n.base)}]%
  \protect\node[fill=cbcol6,shape=rectangle,inner sep=2pt,draw](n){\small #1};}

\usepackage{cleveref}
\crefname{section}{\S}{\S\S}
\Crefname{section}{\S}{\S\S}
\crefname{table}{Tab.}{}
\crefname{figure}{Fig.}{}
\crefname{algorithm}{Alg.}{}
\crefname{equation}{eq.}{}
\crefname{appendix}{App.}{}
\crefformat{section}{\S#2#1#3}  %

\usepackage{todonotes}

\newcommand{\softmax}{\mathrm{softmax}\xspace}
\newcommand{\yy}{\boldsymbol{y}}
\newcommand{\xx}{\boldsymbol{x}}
\newcommand{\cc}{\mathbf{c}}
\newcommand{\al}{\boldsymbol{a}}
\newcommand{\ee}{\mathbf{e}}

\newcommand{\NN}{\mathbf{f}} %
\newcommand{\dech}{\mathbf{h}^{\textit{(dec)}}}
\newcommand{\ench}{\mathbf{h}^{\textit{(enc)}}}
\newcommand{\dece}{\mathbf{e}^{\textit{(dec)}}}
\newcommand{\ence}{\mathbf{e}^{\textit{(enc)}}}
\newcommand{\decb}{\mathbf{b}^{\textit{(dec)}}}

\newcommand{\decU}{\mathbf{U}^{\textit{(dec)}}}

\newcommand{\decV}{\mathbf{V}^{\textit{(dec)}}}

\newcommand{\enchr}{\overrightarrow{\mathbf{h}}^{\textit{(enc)}}}
\newcommand{\encbr}{\overrightarrow{\mathbf{b}}^{\textit{(enc)}}}
\newcommand{\encUr}{\overrightarrow{\mathbf{U}}^{\textit{(enc)}}}
\newcommand{\encVr}{\overrightarrow{\mathbf{V}}^{\textit{(enc)}}}
\newcommand{\enchl}{\overleftarrow{\mathbf{h}}^{\textit{(enc)}}}

\newcommand{\calA}{\mathcal{A}}

\newcommand{\Sigmay}{\Sigma_{\texttt{y}}}
\newcommand{\Sigmax}{\Sigma_{\texttt{x}}}

\newcommand{\defn}[1]{\textbf{#1}}

\newcommand{\softDep}{\numberingBlueB{1}\xspace}
\newcommand{\softDepU}{\numberingBlueB{U}\xspace}
\newcommand{\hardDep}{\numberingRedB{2}\xspace}
\newcommand{\softInd}{\numberingYellowB{3}\xspace}
\newcommand{\hardInd}{\numberingGreenB{4}\xspace}
\newcommand{\hardIndR}{\numberingGreenB{R}\xspace}
\newcommand{\mono}{\circled{M}\xspace}

\newcommand{\GtP}{\numberingPurpleB{G}\xspace}
\newcommand{\trans}{\numberingOrangeB{T}\xspace}
\newcommand{\morphinf}{\numberingGrayB{I}\xspace}
\newcommand{\correct}{\ding{51}}
\newcommand{\incorrect}{\ding{55}}
\usepackage[safe]{tipa}

\usepackage{inconsolata}

\aclfinalcopy %

\title{Hard Non-Monotonic Attention for Character-Level Transduction}

\author  
  {
	\begin{tabular}{cccc}
	Shijie Wu\raise1.0ex\hbox{\normalfont\normalsize \textschwa} & Pamela Shapiro\raise1.0ex\hbox{\normalfont\normalsize \textschwa} & Ryan Cotterell\raise1.0ex\hbox{\normalfont\normalsize \textschwa}\raise1.0ex\hbox{\normalfont \normalsize ,\textipa{H}}
	\end{tabular}
	\\
    \raise1.0ex\hbox{\normalsize \textschwa}Department of Computer Science, Johns Hopkins University, Baltimore, USA \\
    \raise1.0ex\hbox{\normalsize \textipa{H}}The Computer Laboratory, University of Cambridge, Cambridge, UK\\
	{\tt{\{shijie.wu, pshapiro, ryan.cotterell\}@jhu.edu}}
}

\date{}

\begin{document}
\maketitle
\begin{abstract}
  Character-level string-to-string transduction is an important component of
  various NLP tasks. The goal is to map an input string to an output string,
  where the strings may be of different lengths and have characters taken from
  different alphabets. Recent approaches have used sequence-to-sequence models
  with an attention mechanism to learn which parts of the input string the model
  should focus on during the generation of the output string. Both soft
  attention and hard monotonic attention have been used, but hard non-monotonic
  attention has only been used in other sequence modeling tasks such as image
  captioning \cite{DBLP:conf/icml/XuBKCCSZB15}, and has required a stochastic
  approximation to compute the gradient. In this work, we introduce an exact,
  polynomial-time algorithm for marginalizing over the exponential number of
  non-monotonic alignments between two strings, showing that hard attention
  models can be viewed as neural reparameterizations of the classical IBM Model
  1. We compare soft and hard non-monotonic attention experimentally and find
  that the exact algorithm significantly improves performance over the
  stochastic approximation and outperforms soft attention.
  Code is available at \url{https://github.com/shijie-wu/neural-transducer}.
\end{abstract}

\section{Introduction}
Many natural language tasks are expressible as string-to-string transductions operating at the character level. Probability
models with recurrent neural parameterizations currently hold the
state of the art on many such tasks. On those
string-to-string transduction tasks that involve a mapping between two
strings of different lengths, it is often necessary to resolve which
input symbols are related to which output symbols. As an example,
consider the task of transliterating a Russian word into the Latin
alphabet. In many cases, there exists a one-to-two mapping between
Cyrillic and Latin letters: in
\textit{\foreignlanguage{russian}{Хурщёв}} (\textit{Khrushchev}), the Russian
\foreignlanguage{russian}{Х} can be considered to generate the Latin
letters \textit{Kh}. Supervision is rarely, if ever, provided at the
level of character-to-character alignments---this is the
problem that attention seeks to solve in neural models.

With the rise of recurrent neural networks, this problem has been handled with ``soft'' \defn{attention} rather than traditional hard \defn{alignment}. Attention
\cite{bahdanau+al-2014-nmt} is often described as ``soft,'' as it does
not clearly associate a single
input symbol with each output symbol, but rather offers a fuzzy notion of what input
symbols may be responsible for which symbols in the output. In contrast, an alignment
directly associates a given input symbol with a given output
symbol. To express uncertainty, practitioners often place a distribution
over the exponential number of hard non-monotonic alignments, just as a probabilistic parser
places a distribution over an exponential number of
trees. The goal, then, is to learn the parameters of this
distribution over all non-monotonic alignments through backpropagation. Incorporating hard alignment into
probabilistic transduction models dates back much farther in the
NLP literature; arguably, originating with the seminal paper by
\newcite{brown1993mathematics}. Some neural approaches have moved back towards this approach of a more rigid alignment, referring to it as ``hard attention.'' We will refer to this as ``hard attention'' and to more classical approaches as ``alignment.''

This paper offers two insights into the usage of hard alignment. First, we
derive a dynamic program for the exact computation of the likelihood
in a neural model with latent hard alignment: Previous work has used a
stochastic algorithm to approximately sum over the exponential number
of alignments between strings. In so doing, we go on to relate neural
hard alignment models to the classical IBM Model 1 for alignment in
machine translation.  Second, we provide an experimental comparison
that  indicates hard attention models outperform soft attention
models on three character-level string-to-string transduction tasks:
grapheme-to-phoneme conversion, named-entity transliteration and
morphological inflection.

\begin{figure}
  \setlength{\belowcaptionskip}{-10pt}
  \includegraphics[width=\columnwidth]{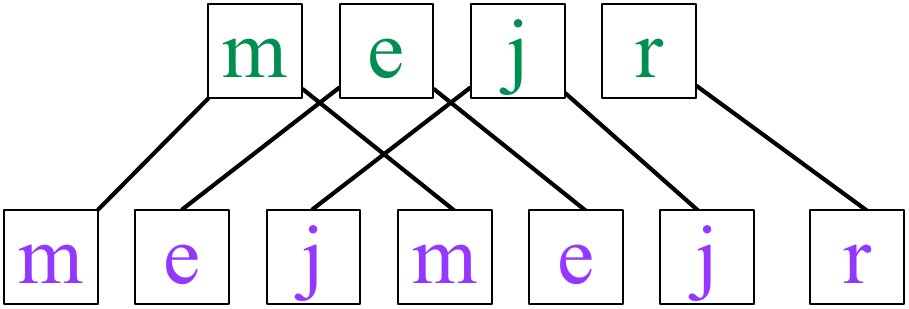}
  \caption{Example of a non-monotonic character-level transduction from the Micronesian language of Pingelapese.
    The infinitive \textit{mejr} is mapped through a reduplicative process to its gerund \textit{mejmejr} \cite{rehg1981rehg}. 
    Each input character is drawn in green and each output character is drawn in purple,
    connected with a line to the corresponding input character.
  }
  \label{fig:pingelapese}
\end{figure}

\section{Non-Monotonic Transduction}\label{sec:non-monotonic}
This paper presents a novel, neural, probabilistic latent-variable model
for \textbf{non-monotonic} transduction.
As a concrete example of a
non-monotonic transduction, consider the mapping of a Pingelapese
infinitive to its gerund, as shown in \cref{fig:pingelapese}. %
The
mapping requires us to generate the output string left-to-right, bouncing around the input string out-of-order to determine the characters to transduce from. As the non-monotonic alignment is the latent
variable, we will face a combinatorial problem: summing over all
non-monotonic alignments. The algorithmic contribution of this paper
is the derivation of a simple dynamic program for computing this
sum in polynomial time that still allows for very rich recurrent neural
featurization of the model.  With respect to the literature, our paper
represents the first instance of exact marginalization for a neural
transducer with hard non-monotonic alignment; previous methods, such
as \newcite{rastogi-cotterell-eisner:2016:N16-1} and
\newcite{aharoni-goldberg:2017:Long}, are exclusively monotonic.

Non-monotonic methods dominate character-level transduction. Indeed,
the state of art in classic character-level NLP tasks such as
grapheme-to-phoneme conversion \cite{DBLP:conf/interspeech/YaoZ15},
transliteration \cite{rosca2016sequence} and morphological inflection
generation \cite{kann-schutze:2016:P16-2} is held by the soft non-monotonic
method of \newcite{bahdanau+al-2014-nmt}. Even though non-monotonicity
is more common in word-level tasks, it also exists in
character-level transduction tasks, as evidenced by our example in
\cref{fig:pingelapese} and the superior performance of non-monotonic
methods. %
Our error analysis in \cref{sec:error-analysis} sheds some light on why
non-monotonic methods are the state of the art in a seemingly monotonic task.

\paragraph{A Note on the Character-level Focus.} A natural question
at this point is why we are not experimenting with word-level
transduction tasks, such as machine translation. As we show in the
\cref{sec:algorithm-analysis} our method is often an order of
magnitude slower than standard soft attention.
Thus,
the exact marginalization scheme is practically unworkable for machine
translation; we discuss future extensions for machine translation in
\cref{sec:future-work}. However, the slow-down is no problem
for character-level tasks and we show empirical
gains in \cref{sec:experiments}.%

\section{Hard Non-Monotonic Alignment}\label{sec:hard-alignment}

\subsection{The Latent-Variable Model}
An alphabet is a %
finite, non-empty set.  Given two alphabets $\Sigmax =
\{x_1, \ldots, x_{|\Sigmax|}\}$ and $\Sigmay = \{ y_1, \ldots,
y_{|\Sigmay|} \}$, probabilistic approaches to the problem attempt to
estimate a probability distribution $p(\yy \mid \xx)$ where $\yy \in
\Sigmay^*$ and $\xx \in \Sigmax^*$. Foreshadowing, we will define the
parameters of $p$ to be, in part, the parameters of a recurrent neural network, in
line with the state-of-the-art models.
We define the set $\calA =\{1, \ldots, |\xx|\}^{|\yy|}$, which has an
interpretation as the set of all (potentially non-monotonic)
alignments from $\xx$ to $\yy$ with the implicit constraint that each output symbol $y_i$ aligns to exactly one symbol in $\xx \in \Sigmax^*$. In other words,
$\mathcal{A}$ is the set of all many-to-one alignments between $\xx$ and $\yy$ where many
may be as few as zero. We
remark that $|\calA| = |\xx|^{|\yy|}$, which is exponentially large in the
length of the target string $\yy$. For an $\al \in \calA$,
$A_i = a_i$ refers to the event that $y_i$, the $i^\text{th}$ component of $\yy$,
is aligned to $x_{a_i}$, the ${a_i}^\text{th}$ component of $\xx$.

We define a probability distribution over output strings $\yy$
conditioned on an input string $\xx$ where we marginalize
out \emph{unobserved} alignments $\al$:
\begin{subequations}
\begin{align}
  &p(\yy \mid \xx) = \sum_{\al \in \calA} p(\yy, \al \mid \xx)  \\
  &= \underbrace{\sum_{\al \in \calA} \prod_{i=1}^{|\yy|} p(y_i \mid a_i, \yy_{< i}, \xx)\, p(a_i \mid \yy_{< i}, \xx)}_{\textit{exponential number of terms}}\label{eq:exp}  \\  
  &= \underbrace{\prod_{i=1}^{|\yy|} \sum_{a_i=1}^{|\xx|} p(y_i  \mid a_i, \yy_{< i}, \xx)\,p(a_i \mid \yy_{< i}, \xx )}_{\textit{polynomial number of terms}}  \label{eq:poly}  \\
  &= \prod_{i=1}^{|\yy|} \sum_{j=1}^{|\xx|} \alpha_{j}(i) \, p(y_i  \mid j, \yy_{< i},  \xx) \label{eq:final}
\end{align}
\end{subequations}
where the transition from (1a) to (1b) follows from the independence assumption and we define $\alpha_j(i) = p(a_i \mid \yy_{< i}, \xx )$ and substitute $j=a_i$, in order to better notationally
compare our model to that of \newcite{bahdanau+al-2014-nmt} in \cref{sec:soft-attention}. 
Each distribution $p(y_i  \mid j, \yy_{< i}, \xx)$ in the definition
of the model has a clean interpretation as
a distribution over the output vocabulary $\Sigmay$, given
an input string $\xx \in \Sigmax^*$, where $y_i$ is aligned
to $x_j$. Thus, one way of thinking about this hard alignment model is
as a product of mixture models, one mixture at each
step, with mixing coefficients $\alpha_j(i)$.\looseness=-1

\paragraph{A Note on $\eos$.}
We suppressed $\eos$ in the autoregressive models above for brevity, e.g., $p(y_i \mid j, \yy_{< i}, \xx)$ is a distribution over $\Sigmay \cup \{\eos\}$, in order for $p(\yy \mid \xx)$ to be a probability distribution.\looseness=-1

\begin{figure*}
\centering
\begin{tikzpicture}
  \tikzstyle{main}=[circle, inner sep = 0cm, minimum size = 10mm, thick, draw =black!80, node distance = 10mm]
  \tikzstyle{blank}=[circle, inner sep = 0cm, minimum size = 10mm, thick, node distance = 11mm]
\tikzstyle{deterministic}=[diamond, inner sep = 0cm, minimum size = 10mm, thick, draw =black!80, node distance = 10mm]
\tikzstyle{connect}=[-latex, thick]
\tikzstyle{box}=[rectangle, draw=black!100]

\node[blank] (X1) [] {};
\node[blank] (X2) [right=of X1] {};
\node[main,fill=black!10] (X3) [right = 0.05cm of X2] {$\xx$};
  \node[main] (A1) [below=of X1] {$a_1$};
  \node[main] (A2) [right=of A1,below=of X2] {$a_2$};
  \node[main] (A3) [right=of A2] {$a_3$};
  \node[main] (A4) [right=of A3] {$a_4$};

  \node[deterministic] (H1) [below=of A1] {$\dech_1$};
  \node[deterministic] (H2) [right=of H1,below=of A2] {$\dech_2$};
  \node[deterministic] (H3) [right=of H2,below=of A3] {$\dech_3$};
  \node[deterministic] (H4) [right=of H3,below=of A4] {$\dech_4$};

  \node[main,fill=black!10] (Y1) [below=of H1] {$y_1$};
  \node[main,fill=black!10] (Y2) [right=of Y1,below=of H2] {$y_2$};
  \node[main,fill=black!10] (Y3) [right=of Y2,below=of H3] {$y_3$};
  \node[main,fill=black!10] (Y4) [right=of Y3,below=of H4] {$y_4$};

  \path (H1) edge [connect] (Y1);
  \path (Y1) edge [connect] (H2);
  \path (H1) edge [connect] (H2);
  \path (H1) edge [connect] (A2);
  \path (A1) edge [connect,bend left] (Y1);
  
  \path (H2) edge [connect] (Y2);
  \path (Y2) edge [connect] (H3);
  \path (H2) edge [connect] (H3);
  \path (A2) edge [connect,bend left] (Y2);
  
  \path (H3) edge [connect] (Y3);
  \path (A3) edge [connect,bend left] (Y3);
  \path (H2) edge [connect] (A3);
  \path (H3) edge [connect] (A4);
  \path (H3) edge [connect] (H4);
  \path (Y3) edge [connect] (H4);
  \path (A4) edge [connect,bend left] (Y4);
  \path (H4) edge [connect] (Y4);
  
  \path[dashed] (A1) edge [connect] (H1);
  \path[dashed] (A2) edge [connect] (H2);
  \path[dashed] (A3) edge [connect] (H3);
  \path[dashed] (A4) edge [connect] (H4);

  \path (X3) edge [connect] (A1);
  \path (X3) edge [connect] (A2);
  \path (X3) edge [connect] (A3);
  \path (X3) edge [connect] (A4);

\end{tikzpicture}
\caption{Our hard-attention model \emph{without input feeding} viewed as a graphical model.
Note that the circular nodes are random variables and
the diamond nodes are deterministic variables ($\dech_i$ is first discussed in
\cref{sec:decoder}). The independence assumption between the alignments $a_i$ when the $y_i$ are \emph{observed} becomes clear. Note that we have omitted
arcs from $\xx$ to $y_1$, $y_2$, $y_3$, and $y_4$ for clarity (to avoid crossing arcs).
We alert the reader that the dashed edges show the additional dependencies added
in the \emph{input feeding version}, as discussed in \cref{sec:input-feeding}.
Once we add these in, the $a_i$ are no longer independent and break exact
marginalization.
Note the hard-attention model does not enforce an exact one-to-one constraint.
Each source-side word is free to align with many of the target-side words,
independent of context.  In the latent variable model, the $x$ variable is a
vector of source words, and the alignment may be over more than one element of
$x$.}
\label{fig:graphical-model}
\end{figure*}
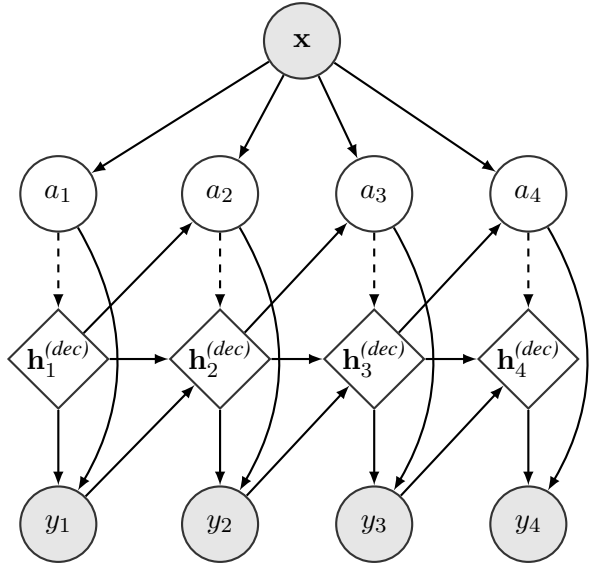

\paragraph{Why Does Dynamic Programming Work?}
Our dynamic program to compute the likelihood, fully specified in \cref{eq:poly}, is quite simple:
The non-monotonic alignments are independent of each other, i.e., $\alpha_j(i)$ is independent
of $\alpha_{j}(i-1)$, %
conditioned on the observed sequence $\yy$.
This means that we can cleverly
rearrange the terms in \cref{eq:exp} using the distributive
property. Were this not the case, we could not do better than
having an exponential number of summands.
This is immediately clear when we view our model as a graphical model, as
in \cref{fig:graphical-model}: There is no active trail
from $a_i$ to $a_k$ where $k > i$, ignoring the dashed lines. Note that
this is no different than the tricks used to achieve exact inference
in $n^\text{th}$-order Markov models---one makes an independence
assumption between the current bit of structure and the previous bits
of structure to allow an efficient algorithm. For a proof
of \cref{eq:exp}--\cref{eq:poly}, one may look in \newcite{brown1993mathematics}. Foreshadowing, we note that
certain parameterizations make use of input feeding \citep[\S3.3]{Luong2015EffectiveAT}, which breaks this independence; see \cref{sec:input-feeding}.\looseness=-1

\paragraph{Relation to IBM Model~1.}
The derivation above is similar to that of the IBM alignment model
1. We remark, however, two key generalizations that will serve our
recurrent neural parameterization well in
\cref{sec:parameterization}. First, traditionally, derivations of IBM
Model~1 omits a
prior over alignments $p(a_i \mid \xx)$, taking it to be
uniform.  Due to this omission, an additional multiplicative constant
$\sfrac{\varepsilon}{|\xx|^{|\yy|}}$ is introduced to ensure the
distribution remains normalized \cite{koehn_2009}. Second, IBM Model 1 does not condition
on previously generated words on the output side. In other words, in
their original model, \newcite{brown1993mathematics} assume that $p(y_i
\mid a_i, \yy_{< i}, \xx) = p(y_i \mid a_i, \xx)$, forsaking
dependence on $\yy_{< i}$. We
note that there is no reason why we need to make this independence
assumption---we will likely want a target-side language model in
transduction. Indeed, subsequent statistical machine
translation systems, e.g., \textsc{Moses} \cite{koehn-EtAl:2007:PosterDemo}, integrate a language
model into the decoder. It is of note that many models in NLP have
made similar independence assumptions, e.g., the emission distribution hidden
Markov models (HMMs) are typically taken to be independent of all
previous emissions \cite{rabiner1989tutorial}. These assumptions are generally not necessary.\looseness=-1

\subsection{Algorithmic Analysis: Time Complexity}\label{sec:algorithm-analysis}
Let us assume that the requisite probability distributions are computable in
${\cal O}(1)$ time and the softmax takes ${\cal O}(|\Sigmay|)$.  Then, by
inspection, the computation of the distribution in \cref{eq:final} is ${\cal
O}\left(|\xx|\cdot |\yy| \cdot |\Sigmay|\right)$, as the sum in \cref{eq:poly}
contains this many terms thanks to the dynamic program that allowed us to
rearrange the sum and the product. While this ``trick'' is well known in the NLP
literature---it dates from the seminal work in statistical machine translation
by \newcite{brown1993mathematics}---it has been forgotten in recent formulations
of hard alignment \cite{DBLP:conf/icml/XuBKCCSZB15}, which use stochastic
approximation to handle the exponential summands. As we will see in
\cref{sec:soft-attention}, we can compute the soft-attention model of
\newcite{bahdanau+al-2014-nmt} in ${\cal O}\left( |\xx| \cdot |\yy| + |\yy|
\cdot |\Sigmay| \right)$ time. When $\Sigmay$ is large, for example in the case of
machine translation with tens of thousands of $\Sigmay$ at least, we can
ignore $|\xx| \cdot |\yy|$ in soft-attention model, and the exact
marginalization has an extra $|\xx|$-factor compared to soft-attention model. In
practice, \newcite{shi2017speeding} show the bottleneck of an NMT system is the
softmax layer, making the extra $|\xx|$-factor practically cumbersome.\looseness=-1

\section{Recurrent Neural Parameterization}\label{sec:parameterization}
How do we parameterize $p(y_i \mid a_i, \yy_{< i}, \xx)$ and $\alpha_j(i)$ in our hard, non-monotonic
transduction model? We
will use a neural network identical to the one proposed in the attention-based
sequence-to-sequence model of \newcite{Luong2015EffectiveAT} \emph{without
input feeding} (a variant of \newcite{bahdanau+al-2014-nmt}).

\subsection{Encoding the Input}\label{sec:encoding}
All models discussed in this exposition will
make use of the same mechanism for mapping a source
string $\xx \in \Sigmax^*$ into a fixed-length representation
in $\mathbb{R}^{d_h}$. This mapping will take the form
of a bidirectional recurrent neural network encoder, which works
as follows: each element of $\Sigmax$ is mapped to an embedding
vector of length $d_e$ through a mapping: $\ee: \Sigmax \rightarrow \mathbb{R}^{d_e}$.
Now, the RNN folds the following recursion over the string $\xx$
left-to-right:
\begin{align}
  \enchr_j = \tanh\Big( \encUr \,\ence(x_j) &+ \\
  \encVr \, \enchr_{j-1} &+ \encbr \Big) \nonumber
\end{align}
where we fix the $0^\text{th}$ hidden state $\ench_0$ to the zero
vector and the matrices $\encUr \in \mathbb{R}^{d_h \times
  d_e}$, $\encVr \in \mathbb{R}^{d_h \times d_h}$ and the bias
term $\encbr \in \mathbb{R}^{d_h}$ are parameters to be learned.
Performing the same procedure on the reversed
string and using an RNN with different parameters,
we arrive at hidden state vectors $\enchl_j$. The final
hidden states from the encoder are the concatenation
of the two, i.e., $\ench_j = \enchr_j \oplus \enchl_j$,
where $\oplus$ is vector concatenation.\looseness=-1

As has become standard, we will use an extension to this
recursion: we apply the long short-term
memory \cite[LSTM;][]{hochreiter1997long} recursions, rather than those of a vanilla RNN \cite[Elman
network;][]{elman1990finding}.  

\subsection{Parameterization.}\label{sec:alignment-distribution}
Now, we define the alignment distribution
\begin{subequations}
\begin{align}
  \alpha_j(i) &= \dfrac{\exp(e_{ij})}{\sum_{j'=1}^{|\xx|} \exp(e_{ij'})}  \label{eq:weights}  \\
  e_{ij} &= {\dech_{i}}^{\top}\, \mathbf{T}\, \ench_{j} \label{eq:attention}
\end{align}
\end{subequations}
where $\mathbf{T} \in \mathbb{R}^{d_h \times 2d_h}$
and $\dech_i$, the decoder RNN's hidden state, is defined in \cref{sec:decoder}.
Importantly, the alignment distribution $\alpha_j(i)$ at
time step $i$ will \emph{only} depend on the prefix of the output
string $\yy_{< i}$ generated so far. This is clear since the
output-side decoder is a unidirectional RNN.
We also define
\begin{equation}\label{eq:inside-single}
\begin{split}
  \quad p(y_i \mid a_i,& \,\yy_{< i}, \xx) = \\
  &\softmax \left(\mathbf{W} \,\NN(\dech_i, \ench_{a_i} ) \right)_{y_i}
  \end{split}
\end{equation}
The function $\NN$ is a non-linear and vector-valued; one popular choice of
$\NN$ is a multilayer perceptron with parameters to be learned. We define
\begin{equation}\label{eq:final-context}
\begin{split}
\quad   \NN(\dech_i, &\ench_{a_i} ) = \\
  &\tanh \left(\mathbf{S} \,(\dech_i \oplus \ench_{a_i} ) \right)
\end{split}
\end{equation}
where $\mathbf{S} \in \mathbb{R}^{d_s \times 3d_h}$.

\subsection{Updating the hidden state $\dech_i$}\label{sec:decoder}
The hidden state~$\dech_i$ is also updated through the
LSTM recurrences \cite{hochreiter1997long}.
The RNN version of the recurrence mirrors that of the encoder,
\begin{align}\label{eq:decoding}
  \dech_i = \tanh \Big( \decU &\,\dece(y_{i-1})\,+ \\
  &\decV \, \dech_{i-1} + \decb \Big) \nonumber
\end{align}
where $\dece : \Sigmay \rightarrow \mathbb{R}^{d_e}$ produces
an embedding of each of the symbols in the output alphabet.
What is crucial about this RNN, like the $\alpha_j(i)$, is that it only summarizes
the characters decoded so far \emph{independently of the previous
  attention weights}. In other words, the
attention weights at time step $i$ will have no influence from the
attention weights at previous time steps, shown in
\cref{fig:graphical-model}.  This is what allows for dynamic programming.

\section{Transduction with Soft Attention}\label{sec:soft-attention}
In order to contrast it with the hard alignment mechanism
we develop, we here introduce Luong attention
\cite{Luong2015EffectiveAT} for recurrent neural sequence to sequence
models \cite{DBLP:conf/nips/SutskeverVL14}. Note that this model
will also serve as an experimental baseline in \cref{sec:experiments}.

The soft-attention transduction model
defines a distribution over the output $\Sigmay^*$, much like
the hard-attention model, with the following expression:
\begin{equation}
  p(\yy \mid \xx) = p(\eos \mid \yy, \xx) \prod_{i=1}^{|\yy|} p(y_i \mid \yy_{< i}, \xx)
\end{equation}
where we define each conditional distribution as
\begin{equation}\label{eq:inside}
\begin{split}
 p(y_i \mid &\,\yy_{< i}, \xx) = \\
 &\softmax \left(\mathbf{W} \,\NN(\dech_i, \cc_i ) \right)_{y_i}
\end{split}
\end{equation}
over $\Sigmay \cup \{\eos\}$.
We reuse the function $\NN$ in \cref{eq:final-context}. The hidden state $\dech_i$,
as before, is the $i^\text{th}$ state of a target-side language model that summarizes
the prefix of the string decoded so far; this is explained in \cref{sec:decoder}.
And, finally, we define the \textbf{context vector}
\begin{equation}
  \cc_i = \sum_{j=1}^{|\xx|} \alpha_j(i) \, \ench_j \\
\end{equation}
using the same alignment distribution as in \cref{sec:alignment-distribution}. In the context of the soft-attention model, this distribution is referred
to as the \textbf{attention weights}.

Inspection shows that there is only a small difference between
the soft-attention model presented here and our hard non-monotonic attention
model. The difference is where we place the probabilities $\alpha_j(i)$. In the
soft-attention version, we place them
inside the softmax (and the function $\NN$), as in \cref{eq:inside}, and we have a mixture of
the encoder's hidden states, the context vector, that we feed into the model. On the
other hand, if we place them outside the softmax, we have a mixture
of softmaxes, as shown in \cref{eq:poly}. Both models have \emph{identical}
set of parameters.\looseness=-1

\subsection{Input Feeding: What's That?}\label{sec:input-feeding}
The equations in \cref{eq:decoding}, however, are not the only
approach. \textbf{Input-feeding} is another popular approach that is, perhaps, standard at this point
\cite{Luong2015EffectiveAT}.
Input feeding refers to the setting where the architecture
designer additionally feeds the attention weights
into the update for the decoder's hidden state. This yields the recursion
\begin{align}\label{eq:inputfeed}
  \dech_i = \tanh\Big(  &\decU \,(\dece(y_{i-1}) \oplus \bar{\cc}_{i-1})\,+ \nonumber \\ 
  & \decV \, \dech_{i-1} + \decb \Big) 
\end{align}
where $\bar{\cc}_{i-1} = \NN(\dech_{i-1}, \cc_{i-1} )$. Note that this
requires that $\decU \in \mathbb{R}^{d_h \times (d_e + d_s)}$.
This is the architecture discussed in \newcite[\S
  3.1]{bahdanau+al-2014-nmt}. In contrast to the
architecture above, this architecture has attention
weights that do depend on previous attention weights due to the feeding in of the
context vector $\cc_i$.
See \newcite{cohn-EtAl:2016:N16-1} for another attempt
to incorporate structural biases directly into the attention mechanism such that the attention distribution
is influenced by previous attention distributions.

\subsection{Combining Hard Non-Monotonic Attention with Input Feeding}\label{sec:hard-input}
To combine hard attention with input feeding, \newcite{DBLP:conf/icml/XuBKCCSZB15}
derive a variational lower bound on the log-likelihood through Jensen's
inequality:
\begin{subequations}
\begin{align}
  &\log p(\yy \mid \xx) = \log \sum_{\al \in \calA} p(\yy, \al \mid \xx) \\
  &= \log \sum_{\al \in \calA} p(\al \mid \xx)\, p(\yy \mid \xx, \al)\label{eq:errata1} \\
  &\geq \sum_{\al \in \calA} p(\al \mid \xx) \log p(\yy \mid \xx, \al) 
\end{align}
\end{subequations}
Note that we have omitted the dependence of~$p(\al \mid \xx)$ on the
appropriate prefix of $\yy$; this was done for notational simplicity. Using this
bound, \newcite{DBLP:conf/icml/XuBKCCSZB15} derive an efficient approximation to the gradient using
the REINFORCE trick of \newcite{williams1992simple}. This 
sampling-based gradient estimator is then used for learning
but suffers from high variance. 
We compare our dynamic programming version to this
model in \cref{sec:experiments}.

\begin{table*}
  \centering
  \setlength{\tabcolsep}{10pt}
  \begin{adjustbox}{width=2\columnwidth}
\begin{tabular}{lll} \toprule
& \textbf{source} & \textbf{target} \\ \midrule \vspace{.1cm}
\textbf{Grapheme-to-phoneme conversion} &
\texttt{a c t i o n} &
\texttt{AE K SH AH N} \\
\textbf{Named-entity transliteration} &
\texttt{A A C H E N} &
아 헨  \\
\textbf{Morphological inflection} &
\texttt{N AT+ALL SG l i p u k e} &
\texttt{l i p u k k e e l l e} \\ \bottomrule
\end{tabular}
\end{adjustbox}
\caption{Example of source and target string for each task as processed by the model}
\label{table:example}
\end{table*}

\section{Future Work}\label{sec:future-work}
Just as \newcite{brown1993mathematics} started with IBM Model~1 and
built up to richer models, we can do the same. Extensions, resembling
those of IBM Model~2 and the HMM aligner
\cite{Vogel:1996:HWA:993268.993313} that generalize IBM Model 1, are easily
bolted onto our proposed model as well. If we are willing to perform
approximate inference, we may also consider fertility as
found in IBM Model 4.

In order to extend our method to machine translation (MT) in any practical
manner, we require an approximation to the softmax. Given that the softmax is
already the bottleneck of neural MT models \cite{shi2017speeding}, we can not afford ourselves
a ${\cal O}(|\xx|)$ slowdown during training. 
Many methods have been proposed for approximating the softmax
\cite{goodman2001classes, bengio2003neural, gutmann2010noise}. More recently,
\newcite{chen2016strategies} compared methods on neural language
modeling, and \newcite{Grave2017EfficientSA} proposed a GPU-friendly method.\looseness=-1

\section{The Tasks}

The empirical portion of the paper focuses on character-level
string-to-string transduction problems.  We consider three tasks:
\GtP: grapheme-to-phoneme conversion, \trans: named-entity
transliteration, and \morphinf: morphological inflection.  We describe
each briefly in turn and we give an example of a source and target
string for each task in \cref{table:example}.

\paragraph{Grapheme-to-Phoneme Conversion.}
We use the standard grapheme-to-phoneme conversion (G2P) dataset:
the Sphinx-compatible version of CMUDict \cite{CMUDict} and NetTalk
\cite{Sejnowski1987ParallelNT}. G2P transduces a word, a string of graphemes, to
its pronunciation, a string of phonemes. We evaluate with word error rate (WER)
and phoneme error rate (PER) \cite{DBLP:conf/interspeech/YaoZ15}.  PER is equal
to the edit distance divided by the length of the string of phonemes.

\paragraph{Named-Entity Transliteration.}
We use the NEWS 2015 shared task on machine transliteration
\cite{Zhang2015WhitepaperON} as our named-entity transliteration dataset. It
contains 14 language pairs. Transliteration transduces a named entity from its source
language to a target language---in other words, from a string in the source
orthography to a string in the target orthography. We evaluate with word
accuracy in percentage (ACC) and mean F-score (MFS)
\cite{Zhang2015WhitepaperON}.  For completeness, we include the definition of MFS in \cref{appendix:MFS}.

\paragraph{Morphological Inflection.}
We consider the high-resource setting of task 1 in the CoNLL--SIGMORPHON 2017 shared task
\cite{cotterell-conll-sigmorphon2017} as our morphological inflection dataset. It
contains 51 languages in the high-resource setting. Morphological inflection
transduces a lemma (a string of characters) and a morphological tag (a sequence of subtags)
to an inflected form of the word (a string of characters). We evaluate with
word accuracy (ACC) and average edit distance (MLD)
\cite{cotterell-conll-sigmorphon2017}. %

\section{Experiments}\label{sec:experiments}
The goal of the empirical portion of our paper is to perform a controlled study
of the different architectures and approximations discussed
up to this point in the paper. \Cref{sec:architectures} exhibits
the neural architectures we compare and the main
experimental results\footnote{Because we do not have access to the test
set of \trans, we only report development performance.} are in \cref{table:all}.
In \cref{sec:exp-detail}, we present the experimental minutiae, e.g. hyperparameters. In
\cref{sec:findings}, we analyze our experimental findings. 
Finally, in \cref{sec:error-analysis}, we perform error analysis and
visualize the soft attention weight and hard alignment distribution.

\begin{table}
  \centering
  \setlength{\tabcolsep}{2pt}
  \begin{adjustbox}{width=\columnwidth}
\begin{tabular}{llll} \toprule
&& \textbf{soft attention} & \textbf{hard alignment} \\ \midrule \vspace{.6cm}
\multirow{3}{*}{\rotatebox{90}{\textbf{input-fed}}}  &  yes & \softDep \newcite{bahdanau+al-2014-nmt,Luong2015EffectiveAT} & \hardDep \newcite{DBLP:conf/icml/XuBKCCSZB15} \\
 &  no  & \softInd \newcite{Luong2015EffectiveAT} \emph{without input feeding} & \hardInd \emph{This work} \\ \bottomrule
\end{tabular}
\end{adjustbox}
\caption{The 4 architectures considered in the paper.}
\label{table:compare}
\end{table}

\begin{table*}[]
\centering
\begin{adjustbox}{width=2\columnwidth}
\begin{tabular}{l llll llll llll} \toprule
& \multicolumn{4}{l}{Grapheme-to-Phoneme Conversion (\GtP)} & \multicolumn{4}{l}{Named-Entity Transliteration (\trans)} &
\multicolumn{4}{l}{Morphological Inflection (\morphinf)} \\ \cmidrule(lr){2-5}
\cmidrule(lr){6-9} \cmidrule(lr){10-13}
& \multicolumn{2}{l}{\textbf{Small}} & \multicolumn{2}{l}{\textbf{Large}}
& \multicolumn{2}{l}{\textbf{Small}} & \multicolumn{2}{l}{\textbf{Large}}
& \multicolumn{2}{l}{\textbf{Small}} & \multicolumn{2}{l}{\textbf{Large}} \\
\cmidrule(lr){2-3} \cmidrule(lr){4-5} \cmidrule(lr){6-7} \cmidrule(lr){8-9} \cmidrule(lr){10-11} \cmidrule(lr){12-13}
& \textbf{WER} & \textbf{PER} & \textbf{WER} & \textbf{PER} & \textbf{ACC} &
\textbf{MFS} & \textbf{ACC} & \textbf{MFS} & \textbf{ACC} & \textbf{MLD} & \textbf{ACC} & \textbf{MLD} \\ \midrule \vspace{.1cm}
\softDep  & 33.7 & 0.080 & 30.8 & 0.074   & 38.9 & 0.890 & 39.9 & 0.893   & 91.4 & 0.183 & 91.1 & 0.201 \\
\softDepU & 30.6 & 0.074 & 30.4 & 0.073   & 39.8 & 0.891 & 40.3 & 0.894   & 91.0 & 0.185 & 91.0 & 0.212 \\
\hardDep  & 32.3 & 0.079 & 33.1 & 0.081   & 36.3 & 0.881 & 30.8 & 0.837   & 91.0 & 0.193 & 89.3 & 0.322 \\
\softInd  & 30.3 & 0.074 & 28.6 & 0.070   & \textbf{40.1} & 0.891 & 40.5 & 0.894   & 92.0 & 0.163 & 92.2 & 0.166 \\
\hardInd  & \textbf{29.6} & \textbf{0.072} & \textbf{28.2} & \textbf{0.068}   & 39.8 & 0.891 & \textbf{41.1} & 0.894  & \textbf{92.6} & \textbf{0.151} & \textbf{93.6} & \textbf{0.128} \\
\hardIndR & 30.7 & 0.076 & 29.7 & 0.074   & 37.1 & 0.882 & 36.9 & 0.863   & 91.2 & 0.190 & 92.8 & 0.151 \\
\mono     & 33.9 & 0.082 & 29.9 & 0.072   & 38.8 & \textbf{0.959} & 40.1 & \textbf{0.960}   & 91.7 & 0.160 & 92.8 & 0.141 \\
\bottomrule
\end{tabular}
\end{adjustbox}
\caption{Average test performance on \GtP, \trans and \morphinf averaged across
datasets and languages. See \cref{appendix:breakdown} for full breakdown.}
\label{table:all}
\end{table*}

\subsection{The Architectures}\label{sec:architectures}
The four architectures we consider in controlled comparison are:
\softDep: soft attention \emph{with} input feeding, \hardDep:
hard attention \emph{with} input feeding, \softInd: soft
  attention \emph{without} input feeding and \hardInd: \textbf{hard
  attention \emph{without} input feeding} (our system).  They are also shown in
\cref{table:compare}. As a fifth system, we compare
to the monotonic system \mono:
\newcite{aharoni-goldberg:2017:Long}. 
Additionally, we present \softDepU, a variant of \softDep where the
number of parameters is not controlled for, and
\hardIndR, a variant of \hardInd trained using
REINFORCE instead of exact marginalization.

\begin{table}
\centering
\begin{adjustbox}{width=\columnwidth}
\begin{tabular}{llll} \toprule
            & \textbf{Small} & \textbf{Large} & \textbf{Search range}        \\ \midrule \vspace{.1cm}
\textbf{Emb. dim.}   & 100   & 200   & \{50,100,200,300\}  \\
\textbf{Enc. dim.}   & 200   & 400   & \{100,200,400,600\} \\
\textbf{Enc. layer}  & 1     & 2     & \{1,2,3\}           \\
\textbf{Dec. dim.}   & 200   & 400   & \{100,200,400,600\} \\
\textbf{Dec. layer}  & 1     & 1     & \{1,2,3\}           \\
\textbf{Dropout}     & 0.2   & 0.4   & \{0,0.1,0.2,0.4\}   \\
\textbf{\# param.}    & 1.199M& 8.621M& N/A   \\ \bottomrule
\end{tabular}
\end{adjustbox}
\caption{Model hyperparameters and search range}
\label{table:hyper}
\end{table}

\subsection{Experimental Details}\label{sec:exp-detail}

We implement the experiments with PyTorch \cite{paszke2017automatic} and we port
the code of \newcite{aharoni-goldberg:2017:Long} to admit batched training.
Because we did not observe any improvements in preliminary experiments when
decoding with beam search \footnote{Compared to greedy decoding with an average
error rate of 20.1\% and an average edit distance of 0.385, beam search with
beam size 5 gets a slightly better edit distance of 0.381 while hurting the
error rate with 20.2\%.}, all models are decoded greedily.

\paragraph{Data Preparation.}
For \GtP, we sample 5\% and 10\% of the data as development set and test set, respectively. For
\trans, we only run experiments with 11 out of 14 language pairs\footnote{Ar--En,
En--Ba, En--Hi, En--Ja, En--Ka, En--Ko, En--Pe, En--Ta, En--Th, Jn--Jk and Th--En.} because we do not
have access to all the data.

\paragraph{Model Hyperparameters.}
The hyperparameters of all models are in \cref{table:hyper}. The
hyperparameters of the large model are tuned using
the baseline \softInd on selected languages in \morphinf, and the search range is shown in \cref{table:hyper}.
All three tasks have the same two sets of hyperparameters.
To ensure that \softDep has the same number of parameters as the other models,
we decrease $d_s$ in \cref{eq:final-context} while for the rest of the models
$d_s = 3d_h$. Additionally, we use a linear mapping to merge $\dece(y_{i-1})$ and
$\bar{\cc}_{i-1}$ in \cref{eq:inputfeed} instead of concatenation. The output of
the linear mapping has the same dimension as $\dece(y_{i-1})$, ensuring that the
RNN has the same size.

\mono has quite a different architecture: The input of the decoder
RNN is the concatenation of the previously predicted word embedding,
the encoder's hidden state at a specific step, and in the case of
\morphinf, the encoding of the morphological tag. Differing from
\newcite{aharoni-goldberg:2017:Long}, we concatenate all attributes'
embeddings (0 for attributes that are not applicable) and merge them with a
linear mapping. The dimension of the merged vector and attributes vector
are $d_e$. To ensure that it has the same number of
parameters as the rest of the model, we increase the hidden size of
the decoder RNN.

\begin{figure*}
\subfloat{{\includegraphics[width=\columnwidth]{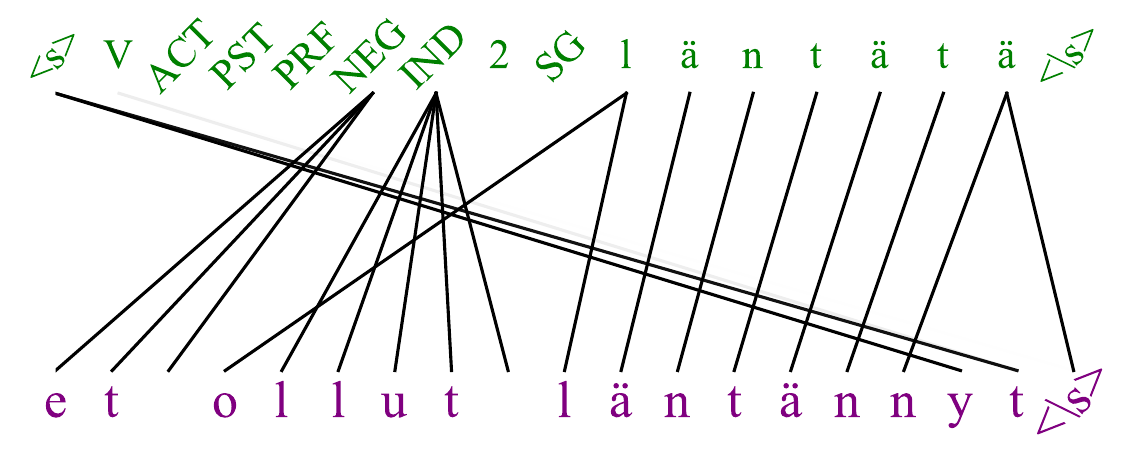} }}
\qquad
\subfloat{{\includegraphics[width=\columnwidth]{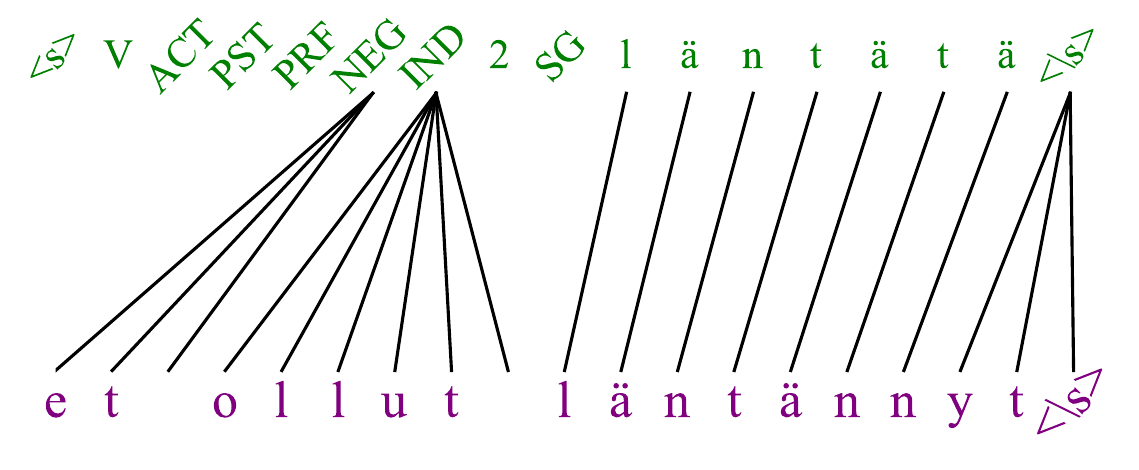}
}}
\caption{Attention-weight (\softInd; left) and alignment distribution (\hardInd;
right) of Finnish in \morphinf. Both models predict correctly. }
\label{fig:vis}%
\end{figure*}

\paragraph{Optimization.}
We train the model with Adam \cite{kingma2014adam} with an initial learning
rate of \(0.001\). We halve the learning rate whenever the development log-likelihood
doesn't improve. We stop after the learning
rate dips to \(1 \times 10^{-5}\). We save all models after each epoch and select the
model with the best development performance. We train the model for at most 50
epochs, though all the experiments stop early. We train on \GtP, \trans, and \morphinf
with batch sizes of  20, 50 and 20, respectively. We notice in the experiments that
the training of \softDep and \softDepU is quite unstable with the large model,
probably because of the longer chain of gradient information flow. We apply
gradient clipping to the large model with maximum gradient norm 5.

\paragraph{REINFORCE.}
In the REINFORCE training of \hardIndR and \hardDep, we sample 2 and 4
positions at each time step for the small and large models,
respectively. The latter is tuned on selected languages in \morphinf
with search range \{2,3,4,5\}. To stabilize the training, we apply a
baseline with a moving average reward and discount factor of 0.9,
similar to \newcite{DBLP:conf/icml/XuBKCCSZB15}.

\subsection{Experimental Findings}\label{sec:findings}
\paragraph{Finding \#1: Effect of Input Feeding.}
By comparing \softInd and \hardInd against \softDep and \hardDep in
\cref{table:all}, we find input feeding hurts performance in all
settings and all tasks. This runs in contrast to the reported results
of \citet{Luong2015EffectiveAT}, but they experiment on machine
translation, rather than character-level transduction. %
This validates our independence assumption about the alignment distribution. 

\paragraph{Finding \#2: Soft Attention vs. Hard Attention.}
Training with REINFORCE hurts the performance of the hard attention model;
compare \softDep and \hardDep (trained with REINFORCE), in \cref{table:all}.  On
the other hand, training with exact marginalization causes the hard attention
model to outperform the soft attention model in nearly all settings; compare
\softInd and \hardInd in \cref{table:all}. This comparison shows that hard
attention outperforms soft attention in character-level string transduction when
trained with exact marginalization.

\paragraph{Finding \#3: Non-monotonicity vs. Monotonicity.}
The monotonic model\mono underperforms compared to non-monotonic models
\softInd in \cref{table:all} except for one setting. It performs slightly worse
on \trans and \GtP due to the many-to-one alignments in the data and the fact
that \newcite{aharoni-goldberg:2017:Long} can only use the hidden vector of the final element of the span in a many-to-one alignment to directly predict the one
target element. The current state-of-the-art systems for character-level string
transduction are non-monotonic models, despite the tasks' seeming monotonicity;
see \cref{sec:error-analysis}.

\begin{table}
\centering
\begin{adjustbox}{width=\columnwidth}
\begin{tabular}{l ll ll ll ll} \toprule
 & \multicolumn{4}{l}{\textbf{NETtalk}} & \multicolumn{4}{l}{\textbf{CMUDict}} \\
  \cmidrule(lr){2-5} \cmidrule(lr){6-9} & \multicolumn{2}{l}{\softInd} & \multicolumn{2}{l}{\hardInd}
  & \multicolumn{2}{l}{\softInd} & \multicolumn{2}{l}{\hardInd} \\
  \cmidrule(lr){2-3} \cmidrule(lr){4-5} \cmidrule(lr){6-7} \cmidrule(lr){8-9}
  & \textbf{\correct} & \textbf{\incorrect} & \textbf{\correct} & \textbf{\incorrect}
  & \textbf{\correct} & \textbf{\incorrect} & \textbf{\correct} & \textbf{\incorrect} \\ \midrule \vspace{.1cm}
  \textbf{Monotonic} & 18742 & 1230 & 18823 & 1172 & 95824 & 17294 & 96176 & 17159 \\
  \textbf{Non-monotonic} & 31 & 5 & 12 & 1 & 158 & 162 & 37 & 66 \\
\bottomrule
\end{tabular}
\end{adjustbox}
\caption{Breakdown of correct and incorrect predictions of monotonic and non-monotonic alignments of \softInd and \hardInd in \GtP, derived from the soft attention weights and the hard alignment distribution}
\label{table:confusion}
\end{table}

\paragraph{Finding \#4: Approximate Hard Attention.}
Given our development of an exact marginalization method for
neural models with hard attention, a natural question to
ask is how much the exact marginalization helps during learning. 
By comparing \hardInd and
\hardIndR in \cref{table:all}, we observe that training with exact
marginalization clearly outperforms training under stochastic approximation
in every setting and on every dataset. We also observe that the exact
marginalization allows faster convergence, since training with REINFORCE is
quite unstable where some runs seemingly to get stuck.

\paragraph{Finding \#5: Controlling for Parameters.}
Input feeding yields a more expressive
model, but also leads to an increase in the number of parameters.
Here, we explore what effect this has on the performance of
the models. In their ablation,
\newcite{Luong2015EffectiveAT} did not control the
  number of parameters when adding input feeding. The total number of
parameters of \softDepU is 1.679M for the small setting and 10.541M
for the large setting, which has 40\% and 22.3\% more parameters than
the controlled setting. By comparing \softDep and \softDepU in
\cref{table:all}, we find that the increase in parameters,
rather than the increase in expressivity explains
the success of input feeding.

\subsection{Visualization and Error Analysis}\label{sec:error-analysis}

We hypothesize that even though the model is non-monotonic, it can learn
monotonic alignment with flexibility if necessary, giving state-of-the-art
results on many seemingly monotonic character-level string transduction tasks. To
show more insights, we compare the best soft attention model (\softInd) against the best
hard alignment model (\hardInd) on \GtP by showing the confusion matrix of each
model in \cref{table:confusion}. 
An alignment is non-monotonic when alignment edges are predicted by the model cross.
There is an edge connecting $x_j$ and $y_i$ if the attention weight or hard
alignment distribution $\alpha_j(i)$ is larger than 0.1. We find that the
better-performing transducers are more monotonic, and most learned alignments
are monotonic. The results indicate that there are a few transductions that are
indeed non-monotonic in the dataset. However, the number is so few that this
does not entirely explain why non-monotonic models outperform the monotonic
models. We speculate this lies in the architecture of
\newcite{aharoni-goldberg:2017:Long}, which does not permit many-to-one
alignments, while monotonic alignment learned by the non-monotonic model is more
flexible. Future work will investigate this.

In \cref{fig:vis}, we visualize the soft attention
weights (\softInd) and the hard alignment distribution (\hardInd) side
by side. We observe that the hard alignment distribution is more
interpretable, with a clear boundary when predicting the prefixes.

\section{Conclusion}
We exhibit an efficient dynamic program for the
exact marginalization of all non-monotonic alignments in a neural
sequence-to-sequence model. We show empirically that the exact
marginalization helps over approximate inference by REINFORCE 
and that models with hard, non-monotonic alignment
outperform those with soft attention.

\paragraph{Acknowledgements.}
RC was supported by an NDSEG fellowship and a Facebook Fellowship.
\bibliography{main}
\bibliographystyle{acl_natbib}

\onecolumn
\appendix
\section{MFS}\label{appendix:MFS}
We consider the following evaluation metrics:   
\begin{subequations}
\begin{align}
\text{LCS}(c_i, r_i) &= \frac{1}{2} (|c_i|+|r_i|-\text{ED}(c_i,r_i)) \\
R_i &= \frac{\text{LCS}(c_i, r_i)}{|r_i|} \\
P_i &= \frac{\text{LCS}(c_i, r_i)}{|c_i|} \\
\text{FS}_i &= 2 \frac{R_i\times P_i}{R_i + P_i}
\end{align}
\end{subequations}
where $r_i$ and $c_i$ is the $i^\text{th}$ reference and prediction and $\text{ED}(c_i, r_i)$ is the
edit distance between $c_i$ and $r_i$.
\section{Full breakdown of experiments}\label{appendix:breakdown}

A full breakdown of \GtP and \trans can be found in \cref{table:full-g2p} and
\cref{table:full-news2015}, respectively. A full breakdown of \morphinf can
be found in \cref{table:full-sigmorphon2017-small} and
\cref{table:full-sigmorphon2017-large}.

\begin{table*}
\centering
\begin{adjustbox}{width=\columnwidth}
\begin{tabular}{l   ll ll ll ll ll ll ll} \toprule &
  \multicolumn{14}{l}{\textbf{Small}} \\
  \cmidrule(lr){2-15} & \multicolumn{2}{l}{\softDep} &
  \multicolumn{2}{l}{\softDepU} & \multicolumn{2}{l}{\hardDep} &
  \multicolumn{2}{l}{\softInd} & \multicolumn{2}{l}{\hardInd} &
  \multicolumn{2}{l}{\hardIndR} & \multicolumn{2}{l}{\mono} \\
  \cmidrule(lr){2-3} \cmidrule(lr){4-5} \cmidrule(lr){6-7} \cmidrule(lr){8-9}
  \cmidrule(lr){10-11} \cmidrule(lr){12-13} \cmidrule(lr){14-15}
& \textbf{WER} & \textbf{PER} & \textbf{WER} & \textbf{PER} & \textbf{WER} &
\textbf{PER} & \textbf{WER} & \textbf{PER} & \textbf{WER} & \textbf{PER} &
\textbf{WER} & \textbf{PER} & \textbf{WER} & \textbf{PER}\\ \midrule \vspace{.1cm}
CMUDict & 36.2 & 0.086 & 31.0 & 0.074 & 35.1 & 0.083 & 30.8 & 0.073 & 30.5 & 0.072 & 31.2 & 0.074 & 32.0 & 0.075 \\
NETtalk & 31.2 & 0.075 & 30.2 & 0.075 & 29.6 & 0.074 & 29.8 & 0.074 & 28.8 & 0.073 & 30.3 & 0.078 & 35.7 & 0.088 \\
\bottomrule
\toprule &
  \multicolumn{14}{l}{\textbf{Large}} \\
  \cmidrule(lr){2-15} & \multicolumn{2}{l}{\softDep} &
  \multicolumn{2}{l}{\softDepU} & \multicolumn{2}{l}{\hardDep} &
  \multicolumn{2}{l}{\softInd} & \multicolumn{2}{l}{\hardInd} &
  \multicolumn{2}{l}{\hardIndR} & \multicolumn{2}{l}{\mono} \\
  \cmidrule(lr){2-3} \cmidrule(lr){4-5} \cmidrule(lr){6-7} \cmidrule(lr){8-9}
  \cmidrule(lr){10-11} \cmidrule(lr){12-13} \cmidrule(lr){14-15}
& \textbf{WER} & \textbf{PER} & \textbf{WER} & \textbf{PER} & \textbf{WER} &
\textbf{PER} & \textbf{WER} & \textbf{PER} & \textbf{WER} & \textbf{PER} &
\textbf{WER} & \textbf{PER} & \textbf{WER} & \textbf{PER}\\ \midrule \vspace{.1cm}
CMUDict & 32.3 & 0.076 & 31.4 & 0.073 & 36.7 & 0.087 & 30.5 & 0.073 & 29.8 & 0.071 & 31.8 & 0.077 & 30.5 & 0.072 \\
NETtalk & 29.3 & 0.071 & 29.4 & 0.072 & 29.5 & 0.075 & 26.8 & 0.068 & 26.6 & 0.066 & 27.7 & 0.071 & 29.3 & 0.072 \\
  \bottomrule
\end{tabular}
\end{adjustbox}
\caption{Full breakdown of G2P}
\label{table:full-g2p}
\end{table*}

\begin{table*}
\centering
\begin{adjustbox}{width=\columnwidth}
\begin{tabular}{l   ll ll ll ll ll ll ll} \toprule &
  \multicolumn{14}{l}{\textbf{Small}} \\
  \cmidrule(lr){2-15} & \multicolumn{2}{l}{\softDep} &
  \multicolumn{2}{l}{\softDepU} & \multicolumn{2}{l}{\hardDep} &
  \multicolumn{2}{l}{\softInd} & \multicolumn{2}{l}{\hardInd} &
  \multicolumn{2}{l}{\hardIndR} & \multicolumn{2}{l}{\mono} \\
  \cmidrule(lr){2-3} \cmidrule(lr){4-5} \cmidrule(lr){6-7} \cmidrule(lr){8-9}
  \cmidrule(lr){10-11} \cmidrule(lr){12-13} \cmidrule(lr){14-15}
& \textbf{ACC} & \textbf{MFS} & \textbf{ACC} & \textbf{MFS} & \textbf{ACC} &
\textbf{MFS} & \textbf{ACC} & \textbf{MFS} & \textbf{ACC} & \textbf{MFS} &
\textbf{ACC} & \textbf{MFS} & \textbf{ACC} & \textbf{MFS}\\ \midrule \vspace{.1cm}
ArEn & 54.9 & 0.954 & 54.8 & 0.953 & 53.9 & 0.954 & 53.5 & 0.951 & 56.6 & 0.954 & 53.9 & 0.950  & 60.1 & 0.980 \\
EnBa & 38.4 & 0.916 & 38.9 & 0.915 & 38.5 & 0.914 & 37.6 & 0.918 & 39.7 & 0.918 & 38.0 & 0.909  & 37.4 & 0.961 \\
EnHi & 42.4 & 0.922 & 44.0 & 0.925 & 40.8 & 0.921 & 46.1 & 0.927 & 43.8 & 0.926 & 43.6 & 0.924  & 43.0 & 0.967 \\
EnJa & 40.5 & 0.871 & 40.6 & 0.868 & 35.4 & 0.853 & 41.2 & 0.872 & 41.3 & 0.872 & 35.2 & 0.852  & 39.4 & 0.952 \\
EnKa & 34.8 & 0.910  & 35.5 & 0.912 & 33.1 & 0.909 & 37.8 & 0.913 & 36.0 & 0.909 & 34.9 & 0.907 & 35.6 & 0.960 \\
EnKo & 52.4 & 0.861 & 51.8 & 0.857 & 49.1 & 0.850  & 54.4 & 0.861 & 54.9 & 0.867 & 48.9 & 0.849 & 47.5 & 0.958 \\
EnPe & 28.3 & 0.899 & 32.6 & 0.908 & 30.8 & 0.903 & 34.8 & 0.911 & 30.5 & 0.901 & 29.6 & 0.898  & 34.7 & 0.964 \\
EnTa & 36.8 & 0.921 & 38.7 & 0.925 & 36.7 & 0.918 & 38.5 & 0.923 & 36.2 & 0.923 & 37.5 & 0.921  & 37.5 & 0.963 \\
EnTh & 42.4 & 0.909 & 42.6 & 0.907 & 37.0 & 0.892 & 42.8 & 0.907 & 42.1 & 0.906 & 37.1 & 0.890  & 40.1 & 0.954 \\
JnJk & 18.1 & 0.717 & 18.1 & 0.717 & 13.4 & 0.693 & 15.3 & 0.706 & 18.1 & 0.716 & 15.5 & 0.705  & 12.7 & 0.933 \\
ThEn & 39.2 & 0.912 & 40.2 & 0.913 & 31.1 & 0.882 & 38.9 & 0.911 & 38.6 & 0.912 & 33.6 & 0.897  & 38.3 & 0.960 \\
\bottomrule
\toprule &
  \multicolumn{14}{l}{\textbf{Large}} \\
  \cmidrule(lr){2-15} & \multicolumn{2}{l}{\softDep} &
  \multicolumn{2}{l}{\softDepU} & \multicolumn{2}{l}{\hardDep} &
  \multicolumn{2}{l}{\softInd} & \multicolumn{2}{l}{\hardInd} &
  \multicolumn{2}{l}{\hardIndR} & \multicolumn{2}{l}{\mono} \\
  \cmidrule(lr){2-3} \cmidrule(lr){4-5} \cmidrule(lr){6-7} \cmidrule(lr){8-9}
  \cmidrule(lr){10-11} \cmidrule(lr){12-13} \cmidrule(lr){14-15}
& \textbf{ACC} & \textbf{MFS} & \textbf{ACC} & \textbf{MFS} & \textbf{ACC} &
\textbf{MFS} & \textbf{ACC} & \textbf{MFS} & \textbf{ACC} & \textbf{MFS} &
\textbf{ACC} & \textbf{MFS} & \textbf{ACC} & \textbf{MFS}\\ \midrule \vspace{.1cm}
ArEn & 52.2 & 0.954 & 54.3 & 0.954 & 0.7 & 0.682 & 55.4 & 0.954 & 55.6 & 0.953 & 57.1 & 0.955  	& 59.7 & 0.979 \\
EnBa & 39.0 & 0.914 & 39.2 & 0.918 & 40.4 & 0.917 & 38.5 & 0.916 & 38.2 & 0.917 & 37.8 & 0.912 	& 37.7 & 0.962 \\
EnHi & 42.0 & 0.923 & 43.0 & 0.926 & 38.9 & 0.914 & 45.7 & 0.929 & 46.1 & 0.928 & 40.7 & 0.916 	& 45.0 & 0.968 \\
EnJa & 40.8 & 0.873 & 40.9 & 0.872 & 37.8 & 0.860  & 41.6 & 0.875 & 40.6 & 0.872 & 39.1 & 0.864	& 41.1 & 0.953 \\
EnKa & 36.2 & 0.913 & 37.9 & 0.914 & 35.0 & 0.909 & 37.5 & 0.913 & 38.6 & 0.915 & 38.2 & 0.913 	& 39.2 & 0.961 \\
EnKo & 53.3 & 0.868 & 53.1 & 0.865 & 50.7 & 0.858 & 53.9 & 0.866 & 55.3 & 0.868 & 49.7 & 0.850 	& 50.1 & 0.961 \\
EnPe & 34.0 & 0.911 & 34.6 & 0.913 & 32.5 & 0.906 & 34.2 & 0.912 & 35.3 & 0.911 & 33.4 & 0.911 	& 34.3 & 0.964 \\
EnTa & 39.1 & 0.925 & 37.6 & 0.922 & 32.5 & 0.901 & 38.5 & 0.925 & 40.2 & 0.927 & 37.1 & 0.919 	& 40.3 & 0.965 \\
EnTh & 43.6 & 0.910  & 43.7 & 0.909 & 32.5 & 0.869 & 43.7 & 0.909 & 43.9 & 0.910  & 40.0 & 0.897& 41.3 & 0.955 \\
JnJk & 18.1 & 0.721 & 18.4 & 0.721 & 0.1 & 0.483 & 17.2 & 0.721 & 17.6 & 0.720  & 0.2 & 0.458 	& 12.6 & 0.934 \\
ThEn & 40.2 & 0.915 & 40.3 & 0.915 & 37.7 & 0.909 & 39.2 & 0.915 & 40.3 & 0.916 & 32.9 & 0.897 	& 39.8 & 0.962 \\
  \bottomrule
\end{tabular}
\end{adjustbox}
\caption{Full breakdown of NEWS2015}
\label{table:full-news2015}
\end{table*}

\begin{table*}
\centering
\begin{adjustbox}{width=\columnwidth}
\begin{tabular}{l   ll ll ll ll ll ll ll} \toprule &
  \multicolumn{14}{l}{\textbf{Small}} \\
  \cmidrule(lr){2-15} & \multicolumn{2}{l}{\softDep} &
  \multicolumn{2}{l}{\softDepU} & \multicolumn{2}{l}{\hardDep} &
  \multicolumn{2}{l}{\softInd} & \multicolumn{2}{l}{\hardInd} &
  \multicolumn{2}{l}{\hardIndR} & \multicolumn{2}{l}{\mono} \\
  \cmidrule(lr){2-3} \cmidrule(lr){4-5} \cmidrule(lr){6-7} \cmidrule(lr){8-9}
  \cmidrule(lr){10-11} \cmidrule(lr){12-13} \cmidrule(lr){14-15}
& \textbf{ACC} & \textbf{MLD} & \textbf{ACC} & \textbf{MLD} & \textbf{ACC} &
\textbf{MLD} & \textbf{ACC} & \textbf{MLD} & \textbf{ACC} & \textbf{MLD} &
\textbf{ACC} & \textbf{MLD} & \textbf{ACC} & \textbf{MLD} \\ \midrule \vspace{.1cm}
albanian-high & 97.2 & 0.048 & 98.5 & 0.023 & 97.9 & 0.043 & 98.1 & 0.031 & 98.1 & 0.045 & 98.5 & 0.029 & 94.5 & 0.150 \\
arabic-high & 89.2 & 0.396 & 79.1 & 0.792 & 91.0 & 0.419 & 90.4 & 0.360 & 91.7 & 0.377 & 90.2 & 0.438 & 89.1 & 0.340 \\
armenian-high & 94.6 & 0.106 & 95.3 & 0.086 & 90.1 & 0.214 & 95.2 & 0.080 & 95.5 & 0.080 & 93.7 & 0.126 & 94.4 & 0.108 \\
basque-high & 100.0 & 0.000 & 100.0 & 0.000 & 100.0 & 0.000 & 99.0 & 0.010 & 97.0 & 0.060 & 100.0 & 0.000 & 95.0 & 0.140 \\
bengali-high & 98.0 & 0.060 & 98.0 & 0.060 & 99.0 & 0.030 & 99.0 & 0.050 & 97.0 & 0.090 & 98.0 & 0.080 & 98.0 & 0.040 \\
bulgarian-high & 88.9 & 0.165 & 88.3 & 0.188 & 96.1 & 0.067 & 94.1 & 0.101 & 93.9 & 0.115 & 95.4 & 0.077 & 96.5 & 0.058 \\
catalan-high & 96.8 & 0.083 & 96.9 & 0.083 & 96.8 & 0.075 & 97.5 & 0.063 & 97.2 & 0.073 & 97.3 & 0.068 & 96.3 & 0.076 \\
czech-high & 90.3 & 0.170 & 92.3 & 0.145 & 90.6 & 0.167 & 91.9 & 0.157 & 92.2 & 0.139 & 78.2 & 0.468 & 91.4 & 0.152 \\
danish-high & 88.9 & 0.166 & 88.9 & 0.174 & 90.1 & 0.151 & 89.1 & 0.170 & 90.2 & 0.148 & 91.4 & 0.132 & 92.6 & 0.118 \\
dutch-high & 93.7 & 0.112 & 94.8 & 0.100 & 95.2 & 0.090 & 95.2 & 0.090 & 94.9 & 0.086 & 94.2 & 0.097 & 94.8 & 0.093 \\
english-high & 96.1 & 0.077 & 96.1 & 0.071 & 90.4 & 0.203 & 96.5 & 0.074 & 95.7 & 0.078 & 95.3 & 0.087 & 96.3 & 0.062 \\
estonian-high & 95.2 & 0.109 & 96.0 & 0.078 & 96.4 & 0.070 & 96.3 & 0.090 & 96.8 & 0.079 & 96.2 & 0.091 & 93.0 & 0.145 \\
faroese-high & 79.9 & 0.420 & 79.2 & 0.390 & 78.4 & 0.449 & 79.5 & 0.413 & 82.9 & 0.365 & 81.6 & 0.392 & 82.5 & 0.333 \\
finnish-high & 86.6 & 0.318 & 81.9 & 0.278 & 86.5 & 0.216 & 88.2 & 0.202 & 90.2 & 0.271 & 84.1 & 0.311 & 86.1 & 0.227 \\
french-high & 84.5 & 0.291 & 85.3 & 0.270 & 85.2 & 0.292 & 83.8 & 0.317 & 85.7 & 0.262 & 82.3 & 0.332 & 86.5 & 0.253 \\
georgian-high & 97.6 & 0.039 & 95.4 & 0.083 & 97.8 & 0.037 & 97.9 & 0.113 & 97.5 & 0.046 & 98.2 & 0.039 & 97.3 & 0.038 \\
german-high & 89.6 & 0.244 & 88.4 & 0.272 & 88.1 & 0.282 & 89.6 & 0.257 & 89.3 & 0.179 & 83.7 & 0.381 & 88.9 & 0.276 \\
haida-high & 97.0 & 0.040 & 98.0 & 0.030 & 98.0 & 0.030 & 99.0 & 0.020 & 97.0 & 0.040 & 98.0 & 0.030 & 92.0 & 0.150 \\
hebrew-high & 99.0 & 0.010 & 98.8 & 0.013 & 99.1 & 0.010 & 97.5 & 0.027 & 97.8 & 0.027 & 97.8 & 0.026 & 98.7 & 0.016 \\
hindi-high & 95.1 & 0.482 & 99.9 & 0.002 & 100.0 & 0.000 & 100.0 & 0.000 & 100.0 & 0.000 & 100.0 & 0.000 & 99.4 & 0.014 \\
hungarian-high & 83.4 & 0.338 & 83.9 & 0.336 & 85.2 & 0.333 & 82.5 & 0.372 & 82.3 & 0.381 & 83.2 & 0.444 & 83.0 & 0.367 \\
icelandic-high & 82.2 & 0.333 & 82.0 & 0.350 & 84.1 & 0.305 & 84.5 & 0.304 & 86.3 & 0.286 & 84.4 & 0.296 & 84.5 & 0.300 \\
irish-high & 87.4 & 0.387 & 84.4 & 0.454 & 89.0 & 0.333 & 87.9 & 0.351 & 90.6 & 0.289 & 88.3 & 0.332 & 88.5 & 0.335 \\
italian-high & 96.1 & 0.101 & 87.2 & 0.251 & 96.0 & 0.099 & 95.5 & 0.111 & 95.7 & 0.106 & 95.5 & 0.105 & 94.6 & 0.120 \\
khaling-high & 99.2 & 0.008 & 98.9 & 0.018 & 98.7 & 0.016 & 98.0 & 0.030 & 98.7 & 0.018 & 98.7 & 0.024 & 98.1 & 0.028 \\
kurmanji-high & 94.2 & 0.123 & 93.7 & 0.101 & 92.3 & 0.184 & 93.2 & 0.126 & 93.8 & 0.098 & 93.1 & 0.143 & 94.0 & 0.074 \\
latin-high & 65.4 & 0.578 & 65.9 & 0.591 & 69.6 & 0.516 & 70.1 & 0.476 & 72.1 & 0.458 & 68.6 & 0.503 & 70.7 & 0.450 \\
latvian-high & 94.7 & 0.084 & 95.3 & 0.071 & 93.7 & 0.104 & 95.1 & 0.090 & 95.5 & 0.081 & 94.2 & 0.101 & 93.6 & 0.100 \\
lithuanian-high & 87.0 & 0.178 & 87.8 & 0.247 & 84.9 & 0.233 & 86.9 & 0.196 & 89.1 & 0.162 & 87.4 & 0.201 & 80.9 & 0.258 \\
lower-sorbian-high & 94.6 & 0.111 & 93.7 & 0.112 & 94.8 & 0.100 & 93.4 & 0.138 & 95.2 & 0.096 & 94.8 & 0.103 & 94.2 & 0.108 \\
macedonian-high & 94.1 & 0.088 & 94.2 & 0.089 & 95.3 & 0.073 & 90.7 & 0.164 & 93.6 & 0.102 & 94.7 & 0.087 & 94.9 & 0.094 \\
navajo-high & 84.9 & 0.446 & 84.6 & 0.461 & 81.2 & 0.468 & 86.2 & 0.332 & 88.5 & 0.268 & 85.1 & 0.356 & 79.8 & 0.450 \\
northern-sami-high & 93.9 & 0.112 & 94.2 & 0.099 & 94.8 & 0.125 & 93.6 & 0.145 & 95.4 & 0.089 & 93.2 & 0.143 & 91.8 & 0.154 \\
norwegian-bokmal-high & 86.4 & 0.220 & 87.6 & 0.293 & 88.4 & 0.193 & 89.7 & 0.172 & 90.0 & 0.158 & 88.2 & 0.198 & 90.9 & 0.156 \\
norwegian-nynorsk-high & 71.8 & 0.454 & 78.1 & 0.363 & 76.5 & 0.392 & 77.9 & 0.378 & 77.4 & 0.379 & 81.0 & 0.324 & 88.4 & 0.197 \\
persian-high & 99.4 & 0.013 & 99.4 & 0.013 & 99.4 & 0.012 & 99.1 & 0.016 & 98.9 & 0.017 & 99.2 & 0.017 & 96.8 & 0.064 \\
polish-high & 89.8 & 0.245 & 86.5 & 0.306 & 82.2 & 0.424 & 90.7 & 0.237 & 89.9 & 0.258 & 88.9 & 0.297 & 90.8 & 0.198 \\
portuguese-high & 98.0 & 0.032 & 96.3 & 0.060 & 96.4 & 0.056 & 98.5 & 0.034 & 98.9 & 0.024 & 97.9 & 0.036 & 98.0 & 0.034 \\
quechua-high & 99.4 & 0.013 & 97.9 & 0.045 & 98.7 & 0.040 & 98.2 & 0.053 & 99.6 & 0.020 & 98.2 & 0.058 & 96.4 & 0.087 \\
romanian-high & 83.3 & 0.649 & 83.6 & 0.454 & 75.6 & 0.711 & 81.6 & 0.482 & 84.8 & 0.432 & 83.8 & 0.473 & 84.8 & 0.404 \\
russian-high & 89.4 & 0.263 & 86.8 & 0.303 & 71.2 & 0.960 & 90.1 & 0.271 & 90.1 & 0.242 & 87.7 & 0.312 & 89.8 & 0.233 \\
serbo-croatian-high & 89.6 & 0.209 & 86.8 & 0.241 & 89.0 & 0.244 & 87.2 & 0.256 & 89.5 & 0.236 & 89.9 & 0.225 & 91.8 & 0.159 \\
slovak-high & 90.5 & 0.155 & 89.8 & 0.168 & 87.2 & 0.212 & 91.3 & 0.145 & 89.6 & 0.163 & 90.2 & 0.158 & 92.6 & 0.128 \\
slovene-high & 94.7 & 0.101 & 94.1 & 0.110 & 95.3 & 0.094 & 96.3 & 0.070 & 96.2 & 0.080 & 94.9 & 0.114 & 95.9 & 0.068 \\
sorani-high & 89.3 & 0.131 & 90.0 & 0.124 & 89.2 & 0.127 & 87.7 & 0.148 & 88.4 & 0.149 & 88.9 & 0.143 & 84.0 & 0.222 \\
spanish-high & 95.9 & 0.080 & 94.3 & 0.104 & 96.0 & 0.083 & 95.1 & 0.105 & 95.5 & 0.099 & 94.6 & 0.118 & 94.2 & 0.099 \\
swedish-high & 87.1 & 0.212 & 87.2 & 0.214 & 87.8 & 0.208 & 88.9 & 0.167 & 88.7 & 0.185 & 69.6 & 0.880 & 90.2 & 0.165 \\
turkish-high & 97.1 & 0.069 & 96.9 & 0.078 & 95.6 & 0.104 & 96.8 & 0.073 & 95.4 & 0.099 & 95.5 & 0.127 & 93.3 & 0.146 \\
ukrainian-high & 90.4 & 0.171 & 91.1 & 0.157 & 89.0 & 0.188 & 90.3 & 0.168 & 91.8 & 0.141 & 91.6 & 0.139 & 92.9 & 0.111 \\
urdu-high & 99.4 & 0.009 & 99.6 & 0.007 & 99.2 & 0.012 & 99.5 & 0.009 & 99.4 & 0.009 & 99.7 & 0.005 & 98.2 & 0.027 \\
welsh-high & 96.0 & 0.070 & 96.0 & 0.080 & 98.0 & 0.040 & 96.0 & 0.070 & 99.0 & 0.030 & 98.0 & 0.040 & 98.0 & 0.040 \\
\bottomrule
\end{tabular}
\end{adjustbox}
\caption{Full breakdown of SIGMORPHON2017 with small model}
\label{table:full-sigmorphon2017-small}
\end{table*}

\begin{table*}
\centering
\begin{adjustbox}{width=\columnwidth}
\begin{tabular}{l   ll ll ll ll ll ll ll}
\toprule &
\multicolumn{14}{l}{\textbf{Large}} \\
\cmidrule(lr){2-15} & \multicolumn{2}{l}{\softDep} &
\multicolumn{2}{l}{\softDepU} & \multicolumn{2}{l}{\hardDep} &
\multicolumn{2}{l}{\softInd} & \multicolumn{2}{l}{\hardInd} &
\multicolumn{2}{l}{\hardIndR} & \multicolumn{2}{l}{\mono} \\
\cmidrule(lr){2-3} \cmidrule(lr){4-5} \cmidrule(lr){6-7} \cmidrule(lr){8-9}
\cmidrule(lr){10-11} \cmidrule(lr){12-13} \cmidrule(lr){14-15}
& \textbf{ACC} & \textbf{MLD} & \textbf{ACC} & \textbf{MLD} & \textbf{ACC} &
\textbf{MLD} & \textbf{ACC} & \textbf{MLD} & \textbf{ACC} & \textbf{MLD} &
\textbf{ACC} & \textbf{MLD} & \textbf{ACC} & \textbf{MLD} \\ \midrule \vspace{.1cm}
albanian-high & 97.9 & 0.037 & 96.3 & 0.066 & 98.5 & 0.024 & 98.0 & 0.030 & 98.5 & 0.028 & 98.8 & 0.021 & 96.1 & 0.123 \\
arabic-high & 90.4 & 0.496 & 89.8 & 0.397 & 88.4 & 0.456 & 89.8 & 0.358 & 92.3 & 0.352 & 91.9 & 0.437 & 90.8 & 0.267 \\
armenian-high & 94.3 & 0.112 & 95.6 & 0.166 & 94.3 & 0.106 & 94.9 & 0.078 & 95.8 & 0.075 & 94.6 & 0.095 & 93.6 & 0.110 \\
basque-high & 100.0 & 0.000 & 100.0 & 0.000 & 99.0 & 0.030 & 100.0 & 0.000 & 100.0 & 0.000 & 100.0 & 0.000 & 99.0 & 0.020 \\
bengali-high & 99.0 & 0.020 & 99.0 & 0.050 & 99.0 & 0.050 & 99.0 & 0.050 & 99.0 & 0.050 & 95.0 & 0.150 & 98.0 & 0.050 \\
bulgarian-high & 95.0 & 0.079 & 91.9 & 0.129 & 93.6 & 0.095 & 96.5 & 0.059 & 96.8 & 0.052 & 96.5 & 0.061 & 96.8 & 0.057 \\
catalan-high & 95.8 & 0.102 & 96.9 & 0.074 & 97.1 & 0.068 & 96.9 & 0.064 & 97.9 & 0.056 & 97.4 & 0.064 & 96.3 & 0.074 \\
czech-high & 89.2 & 0.184 & 87.4 & 0.209 & 87.8 & 0.241 & 90.7 & 0.176 & 92.7 & 0.133 & 89.7 & 0.188 & 92.0 & 0.140 \\
danish-high & 88.3 & 0.167 & 87.0 & 0.282 & 89.3 & 0.166 & 88.6 & 0.159 & 91.9 & 0.121 & 91.7 & 0.127 & 92.2 & 0.121 \\
dutch-high & 93.9 & 0.109 & 93.3 & 0.120 & 94.2 & 0.107 & 94.6 & 0.099 & 95.7 & 0.082 & 95.8 & 0.075 & 95.6 & 0.078 \\
english-high & 95.9 & 0.082 & 95.7 & 0.239 & 93.1 & 0.151 & 95.5 & 0.081 & 96.3 & 0.069 & 96.0 & 0.078 & 96.4 & 0.055 \\
estonian-high & 95.5 & 0.235 & 96.9 & 0.069 & 96.7 & 0.069 & 96.3 & 0.082 & 97.6 & 0.064 & 96.1 & 0.086 & 92.7 & 0.140 \\
faroese-high & 80.8 & 0.388 & 80.7 & 0.468 & 80.1 & 0.374 & 83.0 & 0.365 & 84.3 & 0.327 & 82.3 & 0.371 & 84.5 & 0.314 \\
finnish-high & 89.7 & 0.323 & 86.5 & 0.196 & 26.2 & 4.562 & 88.5 & 0.212 & 92.2 & 0.137 & 90.4 & 0.169 & 88.0 & 0.197 \\
french-high & 82.2 & 0.329 & 83.3 & 0.316 & 84.0 & 0.305 & 82.9 & 0.335 & 85.5 & 0.274 & 84.7 & 0.296 & 85.6 & 0.273 \\
georgian-high & 97.5 & 0.043 & 96.8 & 0.060 & 96.9 & 0.056 & 96.8 & 0.064 & 98.2 & 0.027 & 98.4 & 0.022 & 98.6 & 0.018 \\
german-high & 87.7 & 0.459 & 82.4 & 0.505 & 76.5 & 0.751 & 87.4 & 0.309 & 91.3 & 0.141 & 88.7 & 0.233 & 89.4 & 0.244 \\
haida-high & 98.0 & 0.030 & 98.0 & 0.030 & 98.0 & 0.030 & 97.0 & 0.040 & 98.0 & 0.030 & 98.0 & 0.030 & 95.0 & 0.100 \\
hebrew-high & 98.5 & 0.019 & 99.1 & 0.011 & 98.5 & 0.017 & 98.6 & 0.015 & 98.4 & 0.018 & 98.7 & 0.016 & 98.5 & 0.018 \\
hindi-high & 100.0 & 0.000 & 99.7 & 0.099 & 99.9 & 0.001 & 99.8 & 0.002 & 99.9 & 0.003 & 100.0 & 0.000 & 99.8 & 0.006 \\
hungarian-high & 82.7 & 0.361 & 82.0 & 0.362 & 81.7 & 0.400 & 79.0 & 0.437 & 84.1 & 0.347 & 82.0 & 0.386 & 84.7 & 0.329 \\
icelandic-high & 82.6 & 0.341 & 84.4 & 0.309 & 82.9 & 0.338 & 84.7 & 0.301 & 88.0 & 0.244 & 87.6 & 0.248 & 87.6 & 0.245 \\
irish-high & 85.8 & 0.377 & 86.4 & 0.391 & 81.8 & 0.512 & 88.7 & 0.423 & 90.5 & 0.254 & 88.3 & 0.344 & 89.6 & 0.304 \\
italian-high & 95.8 & 0.101 & 96.1 & 0.089 & 95.6 & 0.113 & 96.6 & 0.081 & 96.5 & 0.084 & 96.2 & 0.091 & 96.0 & 0.104 \\
khaling-high & 99.4 & 0.006 & 99.2 & 0.009 & 99.4 & 0.006 & 99.3 & 0.009 & 99.5 & 0.005 & 99.4 & 0.009 & 97.8 & 0.031 \\
kurmanji-high & 93.0 & 0.162 & 93.2 & 0.268 & 91.3 & 0.183 & 93.1 & 0.128 & 93.0 & 0.089 & 92.7 & 0.146 & 94.0 & 0.076 \\
latin-high & 71.6 & 0.518 & 69.6 & 0.513 & 70.4 & 0.485 & 78.0 & 0.371 & 78.4 & 0.361 & 76.2 & 0.392 & 74.7 & 0.378 \\
latvian-high & 80.8 & 0.524 & 91.5 & 0.133 & 94.6 & 0.095 & 94.3 & 0.097 & 96.3 & 0.056 & 95.7 & 0.078 & 95.2 & 0.068 \\
lithuanian-high & 86.4 & 0.368 & 85.4 & 0.394 & 87.9 & 0.173 & 89.7 & 0.150 & 90.6 & 0.126 & 90.7 & 0.139 & 89.4 & 0.149 \\
lower-sorbian-high & 93.9 & 0.118 & 93.7 & 0.208 & 95.2 & 0.094 & 95.2 & 0.094 & 95.1 & 0.100 & 96.3 & 0.073 & 95.6 & 0.083 \\
macedonian-high & 90.8 & 0.126 & 92.7 & 0.111 & 94.4 & 0.085 & 93.7 & 0.097 & 95.9 & 0.067 & 94.6 & 0.085 & 93.8 & 0.110 \\
navajo-high & 88.1 & 0.268 & 86.9 & 0.312 & 90.8 & 0.198 & 88.3 & 0.435 & 91.3 & 0.201 & 88.6 & 0.279 & 84.5 & 0.359 \\
northern-sami-high & 94.9 & 0.129 & 95.8 & 0.082 & 95.3 & 0.121 & 95.9 & 0.087 & 97.5 & 0.075 & 96.4 & 0.103 & 95.0 & 0.090 \\
norwegian-bokmal-high & 84.4 & 0.331 & 85.1 & 0.403 & 88.4 & 0.272 & 88.1 & 0.199 & 88.7 & 0.190 & 89.6 & 0.178 & 91.0 & 0.152 \\
norwegian-nynorsk-high & 73.3 & 0.440 & 76.2 & 0.408 & 78.2 & 0.376 & 79.4 & 0.354 & 80.4 & 0.345 & 82.6 & 0.314 & 89.3 & 0.182 \\
persian-high & 99.6 & 0.006 & 99.2 & 0.011 & 99.3 & 0.108 & 99.5 & 0.014 & 99.3 & 0.014 & 99.6 & 0.011 & 96.7 & 0.066 \\
polish-high & 85.3 & 0.406 & 85.7 & 0.369 & 88.5 & 0.282 & 88.4 & 0.366 & 89.7 & 0.248 & 89.2 & 0.251 & 90.2 & 0.193 \\
portuguese-high & 97.4 & 0.041 & 97.4 & 0.042 & 98.2 & 0.034 & 97.9 & 0.037 & 98.3 & 0.032 & 98.9 & 0.023 & 98.8 & 0.028 \\
quechua-high & 98.9 & 0.019 & 97.9 & 0.059 & 97.8 & 0.048 & 98.6 & 0.037 & 98.9 & 0.032 & 98.8 & 0.037 & 97.7 & 0.057 \\
romanian-high & 84.9 & 0.594 & 83.4 & 0.568 & 47.1 & 2.794 & 85.1 & 0.457 & 86.7 & 0.400 & 86.8 & 0.422 & 86.4 & 0.450 \\
russian-high & 88.0 & 0.353 & 85.8 & 0.458 & 86.9 & 0.324 & 89.8 & 0.432 & 90.5 & 0.244 & 90.3 & 0.244 & 91.2 & 0.220 \\
serbo-croatian-high & 84.9 & 0.307 & 85.0 & 0.277 & 88.8 & 0.323 & 88.8 & 0.236 & 90.9 & 0.187 & 82.0 & 0.426 & 91.4 & 0.185 \\
slovak-high & 87.7 & 0.214 & 89.4 & 0.270 & 89.2 & 0.264 & 89.6 & 0.186 & 92.1 & 0.126 & 90.4 & 0.163 & 92.6 & 0.129 \\
slovene-high & 94.6 & 0.096 & 93.2 & 0.222 & 94.6 & 0.102 & 95.2 & 0.078 & 96.1 & 0.073 & 94.9 & 0.097 & 96.4 & 0.063 \\
sorani-high & 88.8 & 0.138 & 88.4 & 0.144 & 88.9 & 0.133 & 89.4 & 0.123 & 90.3 & 0.121 & 90.1 & 0.120 & 85.9 & 0.180 \\
spanish-high & 94.3 & 0.092 & 95.8 & 0.163 & 96.1 & 0.072 & 96.1 & 0.075 & 96.7 & 0.056 & 96.2 & 0.072 & 95.4 & 0.079 \\
swedish-high & 85.7 & 0.237 & 85.0 & 0.329 & 79.4 & 0.529 & 85.9 & 0.235 & 90.9 & 0.157 & 89.6 & 0.170 & 92.2 & 0.132 \\
turkish-high & 95.9 & 0.103 & 94.9 & 0.226 & 94.8 & 0.118 & 92.4 & 0.172 & 97.0 & 0.063 & 94.8 & 0.121 & 94.3 & 0.128 \\
ukrainian-high & 90.4 & 0.170 & 90.3 & 0.161 & 90.8 & 0.148 & 92.5 & 0.126 & 92.9 & 0.116 & 93.1 & 0.114 & 91.4 & 0.144 \\
urdu-high & 99.3 & 0.012 & 99.4 & 0.010 & 99.0 & 0.015 & 99.6 & 0.007 & 99.3 & 0.014 & 99.6 & 0.006 & 99.1 & 0.013 \\
welsh-high & 96.0 & 0.060 & 98.0 & 0.050 & 98.0 & 0.040 & 97.0 & 0.050 & 97.0 & 0.070 & 97.0 & 0.060 & 97.0 & 0.060 \\
\bottomrule
\end{tabular}
\end{adjustbox}
\caption{Full breakdown of SIGMORPHON2017 with large model}
\label{table:full-sigmorphon2017-large}
\end{table*}
\newpage

\end{document}